\documentclass[letterpaper]{article} 
\usepackage[]{aaai23}  
\usepackage{times}  
\usepackage{helvet}  
\usepackage{courier}  
\usepackage[hyphens]{url}  
\usepackage{graphicx} 

\usepackage{multirow}
\usepackage{tabularray}
\usepackage{amsmath}
\usepackage{caption}
\usepackage{subcaption}
\usepackage{color}

\usepackage{natbib}  
\usepackage{caption} 
\usepackage{algorithm}
\usepackage{algorithmic}

\usepackage{newfloat}
\usepackage{listings}
\DeclareCaptionStyle{ruled}{labelfont=normalfont,labelsep=colon,strut=off} 
\floatstyle{ruled}
\newfloat{listing}{tb}{lst}{}
\floatname{listing}{Listing}
\title{FP8-BERT: Post-Training Quantization for Transformer\thanks{Accepted by DCAA@AAAI 2023}}

\author {
    Jianwei Li\thanks{North Carolina State University, Email: \texttt{jli265@ncsu.edu}},
    Tianchi Zhang\thanks{University of Michigan, Email: \texttt{tonyztc@umich.edu}},
    Ian En-Hsu Yen\thanks{Moffett AI, Email: \texttt{ian@moffett.ai}},
    Dongkuan Xu\thanks{North Carolina State University, Email: \texttt{dxu27@ncsu.edu}}
}
\affiliations {
}

\begin{document}

\maketitle

\begin{abstract}

Transformer-based models, such as BERT, have been widely applied in a wide range of natural language processing tasks. However, one inevitable side effect is that they require massive memory storage and inference cost when deployed in production. Quantization is one of the popularized ways to alleviate the cost. However, the previous 8-bit quantization strategy based on INT8 data format either suffers from the degradation of accuracy in a Post-Training Quantization (PTQ) fashion or requires an expensive Quantization-Aware Training (QAT) process. Recently, a new numeric format FP8 (i.e. floating-point of 8-bits) has been proposed and supported in commercial AI computing platforms such as H100. In this paper, we empirically validate the effectiveness of FP8 as a way to do Post-Training Quantization without significant loss of accuracy, with a simple calibration and format conversion process. We adopt the FP8 standard proposed by~\citet{nvidia_release} in our extensive experiments of BERT variants on GLUE and SQuAD v1.1 datasets, and show that PTQ with FP8 can significantly improve the accuracy upon that with INT8, to the extent of the full-precision model. 
\end{abstract}

\section{Introduction}
A number of large-scale neural network (NN) architectures have been proposed in recent years and achieved remarkable performance in a wide range of tasks. However, large-scale models require colossal training memory, enormous storage space, and expensive inference costs, making them hard to deploy in production. For example, GPT-3, leading in many natural language processing tasks, has 175 billion parameters to evaluate~\cite{gpt3}.

Quantization is one of the most popularized ways to reduce the model size and decrease deployment costs. The core operation is to map a set of continuous real-valued numbers into a fixed discrete set of numbers to reduce the number of bits required for neural networks. This way, computation with low-precision formats can be executed with a smaller circuit area, fewer clock cycles, and less energy, making model inference quick and environment-friendly~\cite{energy, area}. 

\begin{figure}[!tb]
\centering
\includegraphics[width=0.45\textwidth]{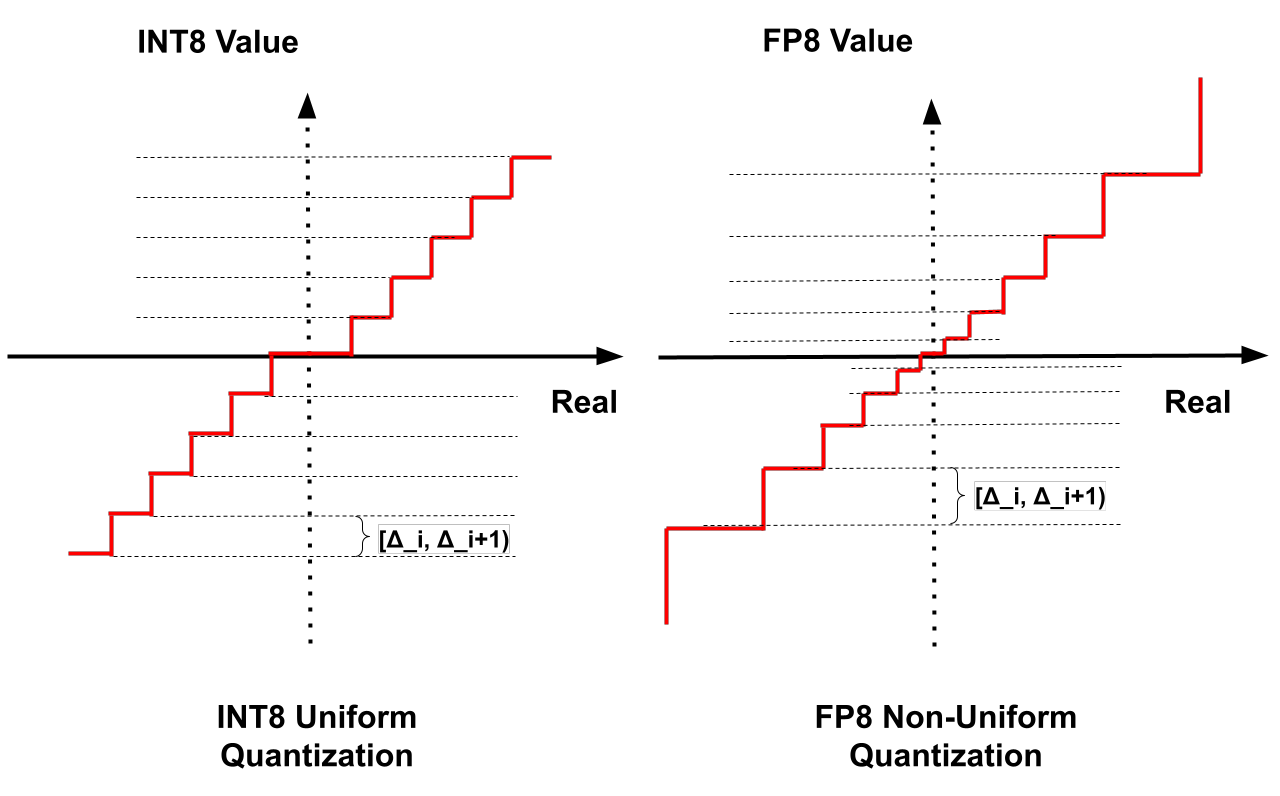}
\caption{INT8 Uniform Quantization v.s. FP8 Non-Uniform Quantization. $[\Delta_{i}, \Delta_{i+1}]$ refers to the discrete quantization interval.}
\label{fig:motivation}
\end{figure}

Moving from full-precision floating-point representations to low-precision fixed integer values represented in eight bits or less is currently the most widely accepted quantization method. For example, \citet{hanint8, gint8, banner} implement quantization with INT8 format and achieves low accuracy loss on CNN-based models. This kind of method primarily trains the model with the full-precision design first and then performs quantization without any finetuning, also known as Post-Training Quantization (PTQ). Therefore, the overhead of PTQ can be negligible. However, it is proved that PTQ with INT8 is unreliable because it may significantly decrease the performance of Transformer-based models~\cite{metalong}. For example, \citet{fp8q_qc} discover that the accuracy of BERT on GLUE datasets significantly drops after quantization. To resolve the problem above, \citet{ibert, q8bert, e8int} finetune the model after quantization to recover the accuracy of the original full-precision model, also known as Quantization Aware Training (QAT). However, its complexity and expensive retraining cost hinder the widespread adoption of QAT.

This paper proposes quantifying the full-precision numbers to floating 8-bit values (FP8) instead of INT8. We delve into the hidden reasons for the failure of PTQ in Transformer-based models and find that there are many outliers in BERT, which is unfriendly for the uniform quantization strategy with INT8. Similarly, \citet{nonuni} shows that, in some trained neural networks, parameters are non-uniformly distributed. Therefore, it is intuitive that non-uniform floating-point 8-bit precision may perform better than uniformly distributed INT8 for quantization. We describe the difference between INT8 uniform quantization strategy and FP8 non-uniform quantization strategy in Figure~\ref{fig:motivation}. 

Previously, FP8 has not been frequently mentioned as a quantization target due to the lack of hardware support. However, Graphcore, AMD, and Qualcomm published their FP8 specification in 2022~\cite{fp8_amd}. At a similar time, NVIDIA, Arm, and Intel released their standard~\cite{fp8_nv}. Both of them include two encodings, E4M3 (1 sign bit, 4 exponents bits, 3 mantissa bits) and E5M2 (1 sign bit, 5 exponents bits, 2 mantissa bits). Moreover, quantization with FP8 becomes realistic as commercial computing platforms with FP8 support release. For example, NVIDIA published RTX 40 Series graphic cards with FP8 support natively~\cite{nvidia_release}. In addition, Habana, Graphcore, and Moffett also announced the plan to support FP8 in the future products~\cite{intel_release, graphcore_release}. 

We describe our contribution as follows:
\begin{itemize}
\item We corroborate the performance of the Post-Training FP8 Quantization method on a wide range of tasks, models, and datasets. 
\item We propose a reliable Post-Training Quantization strategy for BERT and other Transformer-based models that is as simple as the most widely accepted PTQ INT8 Quantization while yielding much closer accuracy to the full-precision model.
\item We provide an empirical FP8 quantization guideline for future research.
\end{itemize}

The rest of the paper is structured as follows. First, we introduce related work in the next section. Then we describe our Post-Training Quantization strategy in the methodology section. Finally, in the experiment section, we compare the results of FP8 quantization strategy and INT8 strategy on Transformer-based and CNN-based models, followed by analysis and conclusion.

\section{Related Work}
Quantization is a method to represent full or higher precision values by a discrete set of numbers to minimize the number of bits required and maximize the accuracy simultaneously. For a machine learning model using single-precision (FP32) format, encoding it with a lower-bit format, like IEEE half-precision format (FP16) or Brain floating point (BF16), can significantly reduce the model size and accelerate model training and inference \cite{tpn, dll, mpt}. 


Quantization with 8-bit integer (INT8) is a popular trend since some edge devices only support integer arithmetic~\citep{arm}. INT8 Quantization has been well explored on many CNN-based models and achieves almost the same accuracy with full-precision versions~\cite{hanint8, gint8, banner}. In parallel, previous works also find that INT8 Quantization with Transformer-based models requires more effort to keep accuracy~\citep{metalong}. \citet{e8int} and \citet{q8bert} quantize BERT with mixed precision and achieve low accuracy loss. \citet{ibert} further enables the BERT model to perform integer-only inference by replacing sensitive operators with specific functions. However, both of these methods require expensive extra model retraining. In this paper, we try to overcome this deficiency with FP8.


Recently, new hardware supporting 8-bit floating numbers (FP8) has been successively released~\cite{nvidia_release, intel_release}, which inspire us to explore 8-bit quantization with FP8.  Early works like \citet{hfp8, fp8_ibm, fp8_intel} has demonstrated FP8's potential ability to maintain accuracy on some CNN-based models compared with FP32. Furthermore,  the parallel work~\cite{fp8q_qc} also indicates that FP8 brings significant improvement of accuracy in the quantization of Transformer-based models, which is consistent with our conclusion.


\section{Methodology\label{sec:methodology}}
This section introduces the FP8 specifications we adopt and then demonstrates the FP8 quantization methods applied in our experiments. Generally, we directly borrow the strategies from INT8 quantization.

\subsection{FP8 Specifications}

The industry has released two FP8 specifications in 2022. One is supported by Graphcore, AMD, and Qualcomm, while NVIDIA, Arm, and Intel propose the other. In this paper, we adopt the standard from~\citet{nvidia_release}, and verify the performance of two kinds of encoding: E4M3 (1 sign bit, 4 exponents bits, 3 mantissa bits) and E5M2 (1 sign bit, 5 exponents bits, 2 mantissa bits).

\subsection{FP8 Simulation\label{sec:conversion}}

Due to the lack of ubiquitous hardware and software supporting FP8 arithmetic, we simulate this low-precision logic with full-precision numbers. Specifically, we implement the C++ library to support the conversion between the FP32 and FP8 numbers. The converting process is rule-based:~(1) the exponent values in two formats should be aligned.~(2) the mantissa value in the casting process should always keep bits with high precision. Finally, we can evaluate FP8 results on general hardware (such as CPU). In addition, we also evaluate the FP8 result on Moffett's next-generation chip (Photon), which natively supports FP8 arithmetic.

\subsection{Non-Uniform FP8 Quantization}

FP8 quantization requires defining the function used to quantize neural network (NN) weights and activations with FP8 format. The core functionality of this function is to map the real values in FP32 precision into a low-precision range. For example,  the integer range for INT8 is [-128, 127]; the float range for FP8 E4M3 is [-448.0, 448.0];  the float range of FP8 E5M2 is [-57334.0, 57334.0]. In this paper, we borrow the idea from INT8 quantization and design Equation~\ref{eq1} to do this work.
\begin{equation} \label{eq1}
Q_{\textit{fp8}}(r) = X_{i}, \textit{if}~\frac{r}{S}~\in~[\delta_{i}, \delta_{i+1})
\end{equation} where $Q_{fp8}$ is the quantization operator, $r$ is the original float number, $S$ is scaling factor, $X_{i}$ denotes the discrete FP8 quantized levels, $\delta_{i}$ represents the FP8 quantized steps. In this paper, we obtain the value of $X_{i}$ by simply casting $\frac{r}{S}$ from FP32 format to FP8 format. Therefore,  we rewrite the function as Equation~\ref{eq2}.
\begin{equation} \label{eq2}
Q_{\textit{fp8}}(r) = \textit{Cast}_{\textit{fp32\_to\_fp8}}(\frac{r}{S}) - Z.
\end{equation} where $Z$ is the offset for zero point. Note that our resulting FP8 quantized values are non-uniformly spaced in 256 8-bit representations because of the nature of floating point numbers. In contrast, the INT8 quantized values are uniformly spaced with the same 
mapping function.

\subsection{Symmetric FP8 Quantization}

According to the definition of Equation~\ref{eq1}, we know the scaling factor $S$ decides the final FP8 quantized values. Therefore, we borrow the idea from INT8 Symmetric Quantization and design Equation~\ref{eq3} to obtain the scaling factor.
\begin{equation} \label{eq3}
S = \frac{\beta - \alpha}{\textit{Max}_{\textit{fp8}}}; \beta = -\alpha
\end{equation} where $[\alpha, \beta]$ is the clipping range of real values. Moreover, the $Z$ in Equation~\ref{eq1} is set to be 0 because of its symmetric property.

\subsection{Clippling Range Calibration}

In order to derive the optimal scaling factor, we may first need to choose the optimal clipping range. Many methods have been proposed to handle this problem. For example, \citet{nvq, McKinstry:2019} uses the min/max value of NN weights or activations to define the clipping range; \citet{migacz:2017} calibrates the clipping range by minimizing KL divergence between the real values and the FP8 quantized values. This paper adopts the most straightforward method: min/max signal. Therefore, the Equation~\ref{eq3} can be rewrite as Equation~\ref{eq4}.
\begin{equation} \label{eq4}
S = \frac{\textit{Max}(|r_{\textit{max}|}, |r_{\textit{min}}|)}{\textit{Max}_{\textit{fp8}}}
\end{equation}

\subsection{Static FP8 Quantization}

At the inference phase, NN activations will change as the input change, so we may generate the scaling factor for each input to reduce quantization error. This method of quantization is known as Dynamic Quantization. However, computing the clipping range dynamically can be expensive. Therefore, we choose static scaling factors for NN activations and weights in this paper.

\subsection{Channelwise and Layerwise FP8 Quantization}

NN weights have multiple dimensions, where the channel dimension is relatively independent. For CNN-based models, each channel corresponds to a number of filters; For Transformer-based models, each channel corresponds to a sequence of token embeddings. The filters and token embeddings from different channels mostly have different ranges of values. Therefore, the granularity of choosing clipping range $[\alpha, \beta]$ for NN weights at each channel is crucial for the final FP8 quantization performance. Currently, Channelwise Quantization is the standard approach used for the NN weights in INT8 quantization, so we adopt the same strategy in FP8 quantization. Regarding the NN activations, we choose the basic Layererwise Quantization strategy.

\subsection{POST Training FP8 Quantization}

For INT8 quantization, previous works mostly require model retraining to recover the accuracy, especially for Transformer-based models. This kind of method is known as Quantization Aware Training. In this paper, our FP8 quantization generally does not require model retraining, also known as Post-Training Quantization.

\subsection{Mixed-precision FP8 Quantization}

Some non-linear operators (such as Softmax, Gelu) and Layernorm are sensitive to Quantization. As a result, it is easy to see significant accuracy degradation when we quantized them. Therefore, we accept half-precision format (BF16) for these special operators in our strategy.

\section{Experiments}

\begin{table*}
\small
\centering
\begin{tblr}{l|c|c|c|c|c|c|c|c}
\hline
Models  & {MNLI-m \\ Acc} & {QNLI \\ Acc} & {QQP \\ Acc} & {MRPC \\ Acc} & {SST-2 \\ Acc} & {COLA \\ Mrr} & {RTE \\ Acc} & {STS-B \\ Spear} & \\ 
\hline
\hline
BERT-base FP32 & 84.58 & 91.4 & 90.91 & 87.6 & 92.43 & 56.3 & 72.3 & 89.0 \\
BERT-base Int8 & 34.39 & 51.62 & 63.16 & 31.62 & 52.64 & 9.0 & 47.29 & 27.6 \\
BERT-base FP8 (E4M3) & 84.63 & 91.58 & 90.96 & 87.2 & 92.32 & 56.0 & 72.2 & 89.0 \\
\hline
\hline
BERT-large FP32 & 86.6 & 92.3 & 91.3 & 89.1 & 93.2 & 60.6 & 74.8 & 90 \\
BERT-large Int8 & 43.25 & 52.12 & 67.33 & 34.52 & 59.33 & 9.2 & 50.12 & 29.32 \\
BERT-large FP8 (E4M3) & 85.9 & 92.2 & 91.3 & 88.9 & 93.3 & 60.0 & 75.1 & 89.8 \\
\hline
\end{tblr}
\caption{BERT: FP32 v.s. INT8 Quantization v.s. FP8 Quantization on GLUE dev set. Note that our quantization results (INT8 and FP8) are both from POST-Training Quantization. We only quantize the general matrix multiply operators(GEMM).}
\label{tab:dev2}
\end{table*}

In this section, we aim to empirically verify the FP8 quantization results in a wide range of tasks, models, and datasets with basic quantization strategies introduced in Section~\ref{sec:methodology}. We also compare our results with INT8 quantization in the same experimental settings. Finally, we explore and analyze the experiment results and the hidden reasons.

\subsection{Baselines and Setup}
We validate the effect of FP8 quantization with BERT-base and BERT-large on GLUE dev datasets for the natural language understanding task. In addition, we also check their quantization results on Squad v1.1 for the question-answering task. Moreover, We do experiments on ResNet18 and ResNet50 with CIFAR10, CIFAR100, and ImageNet datasets for the image classification task.

All the models are first trained and evaluated in FP32 format and then used as our first baseline. After that, the second baseline is derived from INT8 quantization. Note that the two quantization settings are the same as each other. That means the INT8 quantization strategy in the baseline and our FP8 quantization strategy are both Post-Training Quantization, which does not require retraining. In this paper, we only test the quantization results of one encoding of FP8: E4M3.

\subsection{Natural Language Understanding Task}


For Transformer-based models (such as BERT-base and BERT-large), we follow the previous literature~\citep{e8int, q8bert} only to quantize the general matrix multiply operators (GEMM) since they are the bottleneck of execution efficiency of quantized models. In contrast,  we use the half-precision (BF16) to calculate the rest operators (such as LayerNorm, Gelu, and Softmax). This kind of quantization is also known as Mixed-Precision Quantization. Finally, we verify the models' accuracy on the GLUE dev set, and Table~\ref{tab:dev2} describes the FP8 quantization results.  

\subsection{Question Answering Task}

In addition, we also check the FP8 quantization results of Transformer-based models on SQuAD v1.1 
Different from GLUE, SQuAD is more challenging because it is a question-answering task. We show the FP8 quantization results in Table~\ref{tab:dev3}. 

\begin{table}
\small
\centering
\begin{tblr}{l|c}
\hline
Model & squad v1 \\ 
& EM / F1 \\ 
\hline
\hline
BERT-base FP32 & 80.8 / 88.5 \\
BERT-base Int8 & 0.18 / 4.42 \\
BERT-base FP8 (E4M3) & 78.89 / 86.44 \\
\hline
\hline
BERT-large FP32 & 84.1 / 90.9 \\ 
BERT-large Int8 & 0.24 / 6.4 \\
BERT-large FP8 (E4M3) & 82.87 / 88.9 \\
\hline
\end{tblr}
\caption{BERT: FP32 v.s. INT8 Quantization v.s. FP8 Quantization on SQuAD v1.1}
\label{tab:dev3}
\end{table}




\paragraph{Result Analysis.}
Based on the results in Tables~\ref{tab:dev2}-\ref{tab:dev1}, we discover that Post-Trianing FP8 Quantization can significantly improve the accuracy of 8-bit quantized Transformer-based models, even close to the original full-precision models. We conjecture that weights and activations of Transformer-based models consist of a large number of outlier values, which is unfriendly for Post-Training INT8 Quantization. In contrast, FP8 is natively non-uniform and represents a more extensive range of values, which is ideal for the Post-Training Quantization of Transformer-based models. Therefore, our FP8 quantization method is a reliable Post-Training Quantization strategy, and the results of our experiments can provide an empirical guideline for future FP8 quantization research.

\section{Conclusion}
This paper corroborates the performance of Post-Training FP8 Quantization method on a wide range of tasks, models, and datasets. We conclude that FP8 Quantization can produce quantized models with no loss of accuracy for Transformer-based models. We also conclude that FP8 Quantization beats INT8 Quantization in handling outliers due to its non-uniformity of numeric representation.

\bibliography{cite}

\end{document}